\documentclass[conference]{IEEEtran}
\IEEEoverridecommandlockouts
\usepackage{cite}
\usepackage{amsmath,amssymb,amsfonts}
\usepackage{algorithmic}
\usepackage{graphicx}
\usepackage{textcomp}
\usepackage{xcolor}
\def\BibTeX{{\rm B\kern-.05em{\sc i\kern-.025em b}\kern-.08em
    T\kern-.1667em\lower.7ex\hbox{E}\kern-.125emX}}

\makeatletter
\newcommand{\linebreakand}{%
  \end{@IEEEauthorhalign}
  \hfill\mbox{}\par
  \mbox{}\hfill\begin{@IEEEauthorhalign}
}
\makeatother

\begin{document}

\title{CCLAP: CONTROLLABLE CHINESE LANDSCAPE PAINTING GENERATION\\ VIA LATENT DIFFUSION MODEL\thanks{This work is partially supported by National Key R\&D Program of China (No. 2020AAA0104500), and National Natural Science Foundation of China (No. 62176251).}}

\author{\IEEEauthorblockN{~Zhongqi~Wang\IEEEauthorrefmark{1},~Jie~Zhang\IEEEauthorrefmark{2}\IEEEauthorrefmark{3},~Zhilong~Ji\IEEEauthorrefmark{4},~Jinfeng~Bai\IEEEauthorrefmark{4} and~Shiguang~Shan\IEEEauthorrefmark{2}}
\IEEEauthorblockA{\IEEEauthorrefmark{1}School of Computer Science \& Technology, Beijing Institute of Technology, Beijing, China  \\
Email: 1120190892@bit.edu.cn}
\IEEEauthorblockA{\IEEEauthorrefmark{2}Institute of Computing Technology, Chinese Academy of Sciences, Beijing, China }
\IEEEauthorblockA{\IEEEauthorrefmark{3}Institute of Intelligent Computing Technology, Chinese Academy of Sciences, Suzhou, China \\Email: \{zhangjie, sgshan\}@ict.ac.cn}
\IEEEauthorblockA{\IEEEauthorrefmark{4}Tomorrow Advancing Life, Beijing, China  \\Email: jizhilong@tal.com, jfbai.bit@gmail.com}
}

\maketitle

\begin{abstract}
With the development of deep generative models, recent years have seen great success of Chinese landscape painting generation. However, few works focus on controllable Chinese landscape painting generation due to the lack of data and limited modeling capabilities. In this work, we propose a controllable Chinese landscape painting generation method named CCLAP, which can generate painting with specific content and style based on Latent Diffusion Model. Specifically, it consists of two cascaded modules, i.e., content generator and style aggregator. The content generator module guarantees the content of generated paintings specific to the input text. While the style aggregator module is to generate paintings of a style corresponding to a reference image. Moreover, a new dataset of Chinese landscape paintings named CLAP is collected for comprehensive evaluation. Both the qualitative and quantitative results demonstrate that our method achieves state-of-the-art performance, especially in artfully-composed and artistic conception. Codes are available at https://github.com/Robin-WZQ/CCLAP.
\end{abstract}


\section{Introduction}
Chinese landscape painting is a widespread and time-honored oriental art form where artists use a brush and ink to depict the landscape. With the development of deep generative models, it becomes possible that computers can automatically generate Chinese landscape paintings. Recently, many efforts have been devoted to the Chinese landscape painting generation, where we can generally categorize them into three groups, i.e., noise-to-painting \cite{Xue2020EndtoEndCL,Luo2022HighResolutionAA,9903093}, image-to-painting \cite{Zhou2019AnIA,He2018ChipGANAG,Lin2018TransformAS}, and text-to-painting \cite{Yuan2022LearningTG}. 

Noise-to-painting is an unconditional generative task where the model generates paintings seeded from latent space. One of the representative methods is Sketch-And-Paint GAN (SAPGAN) \cite{Xue2020EndtoEndCL}, which successively generates contours and colors to create paintings. To further improve the painting quality, Luo et al. \cite{Luo2022HighResolutionAA} propose a powerful creation system, which includes generating, resizing, and super-resolution, to generate high-resolution and arbitrary-sized paintings. However, although these methods can generate lifelike Chinese landscape paintings, models cannot generate paintings with specific content, which is a vital desideratum to real-world applications. In order to generate paintings with the desired content, some approaches regard the Chinese landscape painting generation as an image-to-painting translation, which takes  photos \cite{He2018ChipGANAG,Lv2019GeneratingCC} or user-defined contours \cite{Zhou2019AnIA,Lin2018TransformAS} as input. However, all these methods highly rely on the quality of the given reference picture. Besides, text-to-painting is another way to fine-grained control over the generated paintings, where models can synthesize paintings from text prompts. Polaca \cite{Yuan2022LearningTG} takes poetry as input and outputs the landscape painting image based on the content of corresponding poetry. Although all the above methods can generate paintings of specific content under user control to some extent, few of them can control the painting to be a given painting style.

It is noticeable that almost all existing works are based on Generative Adversarial Networks (GANs) \cite{Goodfellow2014GenerativeAN}. However, GANs are difficult to train since they always suffer from model collapse and training instabilities \cite{10.5555/3157096.3157346}, leading to unreasonable results for painting generation. Recent works \cite{NEURIPS2020_4c5bcfec,Rombach2021HighResolutionIS} have demonstrated that Diffusion Models (DMs) can get a better generation speaking of diversity, fidelity, and ability of controllable image generation. Thus, it is worth finding out the effectiveness  of diffusion models in the Chinese landscape painting generation. 

\begin{figure}[tbp]
\centerline{\includegraphics[scale=0.37]{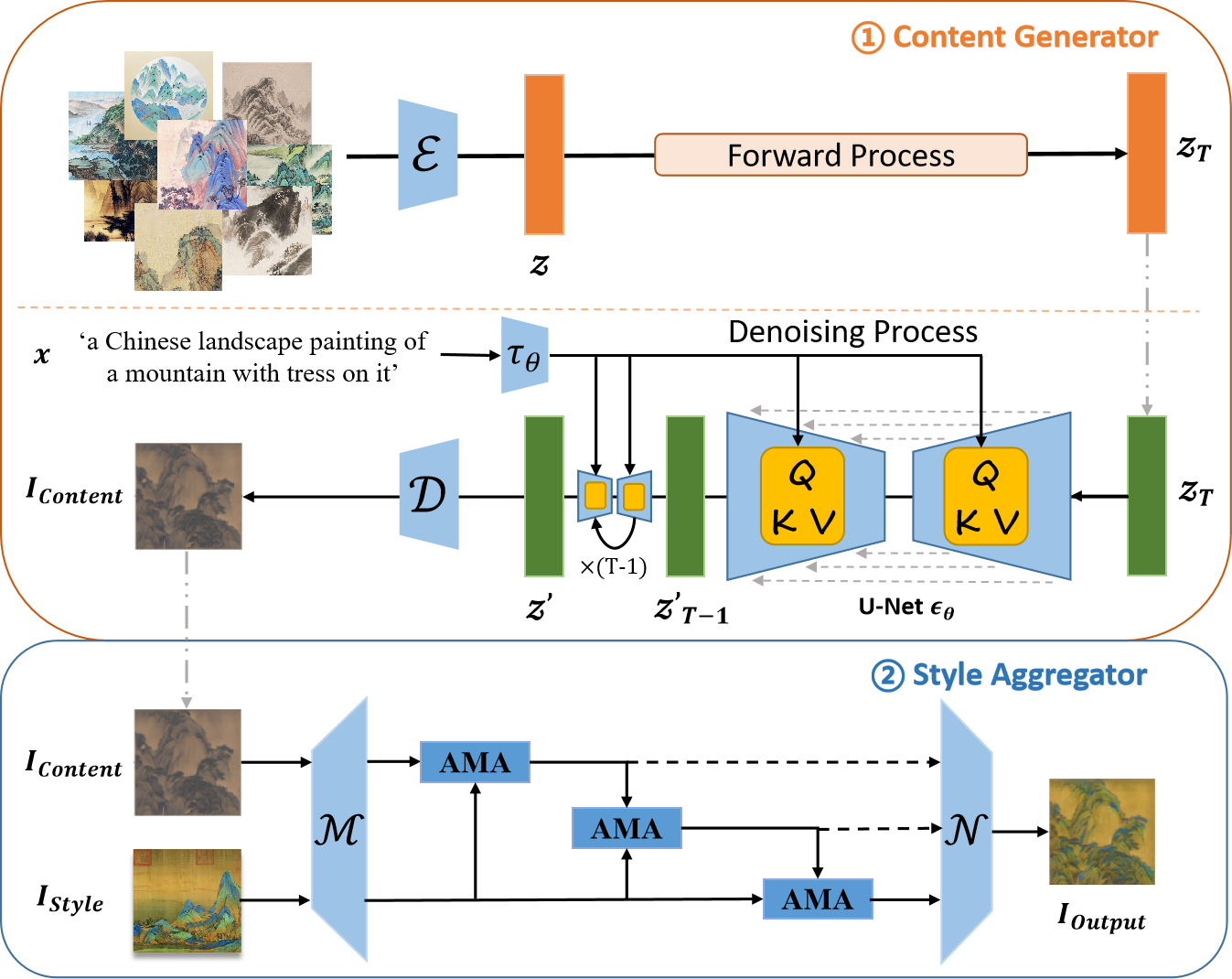}}
\caption{The overview of our CCLAP, which consists of two key parts. 1) A Content Generator \textbf{$\mathcal{C}$} based on latent diffusion model controls the content of the generated painting guided by text prompts $x$. The forward and denoising processes are carried out in the latent space, denoting as an encoder $\mathcal{E}$ and a decoder $\mathcal{D}$, respectively. 2) A Style Aggregator {$\mathcal{S}$} controls the style of the generated painting indicated by image reference, where paintings are encoded and decoded through $\mathcal{M}$ and $\mathcal{N}$, respectively. The content manifold is aligned with the style manifold through three attentional manifold alignment (AMA) blocks step by step.}
\label{fig:model}
\end{figure}

In this work, we propose a method for Controllable Chinese LAndscape Painting generation (\textbf{CCLAP}) to overcome the abovementioned limitations. As shown in Fig.\ref{fig:model}, our method consists of two cascaded modules, i.e., content generator and style aggregator. The generator module generates paintings guided by input texts, which is based on Latent Diffusion Model (LDM) \cite{Rombach2021HighResolutionIS}. Inspired by PAMA \cite{Luo_2022_ACCV}, the style aggregator is designed to generate a specific painting's style indicated by the reference image.  Through these steps, users can get certain landscape paintings in terms of a specific content and style, resulting in a controllable Chinese landscape painting generation. Moreover, since there are no public datasets with both the Chinese landscape painting and the corresponding text, we conduct a text-to-painting dataset CLAP, consisting of 3560 images and the corresponding text descriptions. Both the quantitative and qualitative results on CLAP demonstrate that our CCLAP achieves state-of-the-art performance. 

In summary, we can conclude the main contributions of our method in three aspects.
\begin{itemize}
\item We propose CCLAP for the controllable Chinese landscape painting generation based on the Latent Diffusion Model. To the best of our knowledge, it is the first work for generating landscape paintings according to specific contents and styles.
\end{itemize}
\begin{itemize}
\item Our approach outperforms the state-of-the-art methods in terms of both artfully-composed, artistic conception and Turing test. Results show that the proportion of paintings generated by our model perceived as human creation reaches 61.7$\%$.
\end{itemize}
\begin{itemize}
\item We built a new dataset named CLAP for text-conditional Chinese landscape generation, which may provide a good benchmark for future research in controllable Chinese landscape generation.
\end{itemize}

\section{Related work}
In this section, we review recent works on Chinese landscape painting generation and diffusion model, which are highly related to our approach.

\textbf{Chinese Landscape Painting Generation.} Based on the form of input, we can divide all previous works into three categories, i.e., noise-to-painting, image-to-painting, and text-to-painting. For the first type, paintings are generated by taking random noise as input. Xue \cite{Xue2020EndtoEndCL} proposes SAPGAN, which consists of Sketch-GAN for contour generation and Paint-GAN for contour-based coloring. To reinforce its ability,  Luo et al. \cite{Luo2022HighResolutionAA} propose an automated system where they use StyleGan2 \cite{Karras2019AnalyzingAI} as generation module and ESRGAN \cite{Wang2018ESRGANES} to generate high-resolution and arbitrary-sized paintings. Although their results are lifelike enough, the above models lack the ability to fine-grained control over the generated paintings, limiting their application scope since all paintings are randomly generated.

Another typical approach treats the Chinese landscape painting generation as an image-to-painting translation problem, which takes images like photographs or sketches as inputs. Li et al. \cite{Li2018NeuralAS} propose a style transfer model based on MXDoG operators, but it only captures style information and ignores details. To enrich the local details of the generated painting, Wang et al. \cite{Wang2021AttentionalWN} propose a wavelet transfer model based on the attention mechanism, which can simultaneously model the high-level and low-level information of the paintings. Unlike these two works that paintings are transferred from the photo, Zhou et al. \cite{Zhou2019AnIA} propose a CycleGAN-based model that can color the picture in landscape painting style according to the sketch provided by the user. However, all these methods highly rely on the quality of input photos or sketches, which leads to a massive gap between the generated painting and the real painting.

The most relevant work to our method is Polaca \cite{Yuan2022LearningTG}, which is the first text-to-painting approach in generating Chinese landscape paintings. It is a GAN-based model to generate paintings guided by poetry. Although Polaca can generate paintings of the content specific to the poetry, the painting style is still randomly generated, which is hard be controlled by users. Besides, all the existing Chinese landscape painting generation methods are based on GANs, which are hard to train and suffer from model collapse. To stabilize the training process, many methods have been proposed, including the use of carefully designed network structures \cite{Radford2015UnsupervisedRL} or better loss functions \cite{Arjovsky2017WassersteinG}, but it still leaves a long way to go.

\textbf{Diffusion Model.} Diffusion models are probabilistic deep generative models which showcase impressive generative capabilities in fidelity and diversity of the generated samples. The typical diffusion model \cite{NEURIPS2020_4c5bcfec} has two stages: a forward stage to add Gaussian noise to input data over several steps and a backward step to recover the original input data from the diffused data.

To date, diffusion models have been widely used in generative modeling tasks \cite{Ramesh2022HierarchicalTI,Rombach2021HighResolutionIS}, among which the most closely related task to our method is conditional image synthesis, i.e., text-to-image. Ramesh et al. propose unCLIP (DALLE-2) \cite{Ramesh2022HierarchicalTI}, where a prior model can generate a CLIP-based image embedding based on text conditions, and a diffusion-based decoder can generate images based on the image embedding conditions. To reduce the computational resources for the training diffusion model, Rombach et al. introduce Latent Diffusion Model (LDM) \cite{Rombach2021HighResolutionIS}, which changes the processing object of diffusion process from image space to latent space. It has superior performance in terms of computation cost and inference speed while maintaining high image synthesis quality. Inspired by them, we design a controllable Chinese landscape generation model based on LDM, which can generate paintings of specific content and style to user inputs.

\section{DATASET}

There is yet to be a dataset available for text-guided Chinese landscape painting generation. Thus, we collect a new dataset called CLAP which consists of 3560 paintings of various painting styles with the corresponding text description, to further push the frontier of controllable Chinese landscape painting generation.

\textbf{Collection. }We collect 3560 traditional Chinese landscape paintings from search engines and electronic museums. To guarantee the quality of our dataset, we manually filter out non-landscape artworks and discard low-resolution or unclear paintings.

\begin{figure}[tbp]
    \centerline{\includegraphics[scale=0.42]{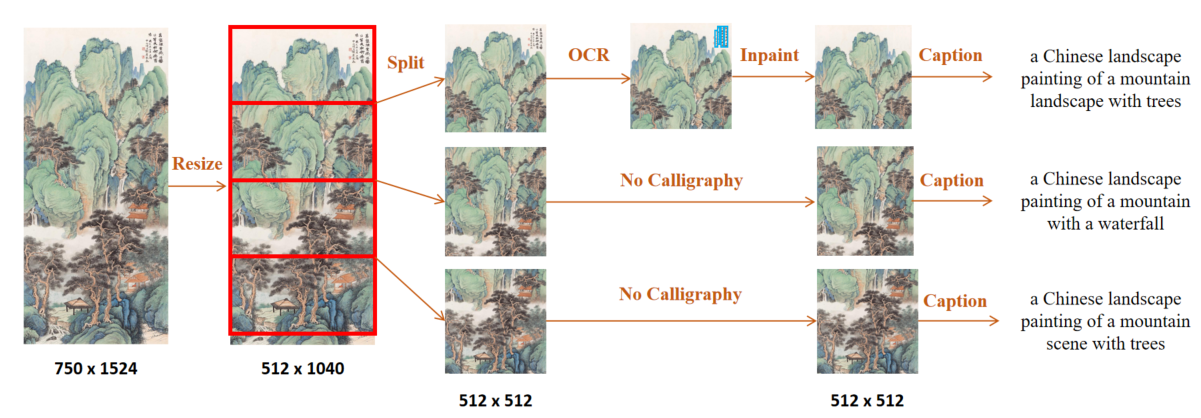}}
    \caption{An example of a painting’s processing steps. A painting of any size is split into multiple paintings with 512$\times$512 resolution and the corresponding text description. }
    \label{fig:dataset}
\end{figure}

\textbf{Preprocessing.} Fig.\ref{fig:dataset}. shows the preprocessing steps. We follow a similar operation described in \cite{Xue2020EndtoEndCL}, where the short side is first scaled to 512 pixels while maintaining the aspect ratio. Then, for paintings with an aspect ratio lower than 1.5, we center-crop them to 512$\times$512 pixels. For others, we use the sliding window to cut into  512$\times$512 pixels with a stepsize of 256 pixels.

\textbf{Cleaning.} We find that most artists prefer to write poems in their works. If we use landscape paintings with ancient poems to train the generative model, the model may generate unreadable results. Thus, it is necessary to remove poems from the original painting. Here we follow a similar method introduced in \cite{Luo2022HighResolutionAA}, which processes these paintings in a coarse-to-fine manner. Specifically, we first adapt a pre-trained text detection model \cite{liao2020real} to find the poem area. Then, we use the fast marching method \cite{Telea2004AnII} for image inpainting.

\textbf{Text-image Pair Generation.} In order to generate Chinese landscape paintings specific to the input texts, we need to obtain the text-image pairs for the whole dataset. Here, instead of manually writing out all captions, we automatically achieve the text descriptions by BLIP \cite{Li2022BLIPBL}, a state-of-art model for image captioning. To emphasize Chinese landscape painting, we replace ``oriental painting’’ or ``drawing’’ and other synonyms with `Chinese landscape painting’’ in each text.

\section{APPROACH}

\subsection{Overview}
We propose CCLAP for controllable Chinese landscape painting generation, which consists of two key parts, i.e. Content Generator $\mathcal{C}$ and Style Aggregator $\mathcal{S}$. The overview of our method is shown in Fig.\ref{fig:model}. Given an input text $x$ and a reference style painting $I_{Style}$, our method try to generate paintings $I_{Output}=\mathcal{S}(\mathcal{C}(x,I_{Style}))$, which can be decomposed into following cascaded objectives:

\begin{equation}
    I_{Content} := \mathcal{C}(\tau_\theta(x))
\end{equation}
\begin{equation}
    I_{Output} := \mathcal{S}(I_{Content},I_{Style})
\end{equation}
Where $\tau_\theta$ is the model that maps text description to an intermediate representation.

\subsection{Content Generator}

Latent Diffusion Model (LDM) \cite{Rombach2021HighResolutionIS} is a recent work which carries out the diffusion process on the low-dimensional latent space. Thanks to the general latent space, it has cheaper training computation cost, and faster inference speed, while maintaining high image synthesis quality. 

Inspired by LDM, we conduct a content generator $\mathcal{C}$ for text guided painting generation, including an encoder $\mathcal{E}$, a decoder $\mathcal{D}$. Specifically, an image $I \in \mathbb{R}^{\mathrm{H}\times\mathrm{W}\times 3 }$ is mapped to a latent vector $z = \mathcal{E}(I)$ through encoder $\mathcal{E}$ , and samples from $p (z)$ can be decoded to an image through $\mathcal{D}$, giving $I_{Content} = \mathcal{D} (z)$. Then, given input text $x$ , a domain specific encoder $\tau_\theta$ projects $x$ into an intermediate representation $\tau_\theta(x)$, which is further fed to the intermediate layers of the UNet via a cross-attention layer implementing Attention($Q,K,V$) = softmax($\frac{QK^T}{\sqrt{d}}$)$\cdot V$, with $Q=W_Q^{(i)}·\phi_i(z_t), K=W_K^{(i)}·\tau_\theta(x),V=W_V^{(i)}·\tau_\theta(x)$. Here, $\phi_i(z_t)$ is a intermediate representation of a  denoising UNet $\epsilon_\theta(z_t,t,\tau_\theta (x)),t=1\dots T$, and $W_Q, W_K, W_V$ are learnable parameters. The corresponding objective can be described as
\begin{equation}
    \mathcal{L}_{C} := \mathbb{E}_{\mathcal{E}(I),x,\epsilon \sim N(0,1),t} [||\epsilon-\epsilon_\theta(z_t,t,\tau_\theta (x))||_2^2]
\end{equation}
where $t$ is uniformly sampled from ${1, . . . , T}$ and $\epsilon$ follows the standard normal distribution.

\begin{figure}[htbp]
    \centerline{\includegraphics[scale=0.365]{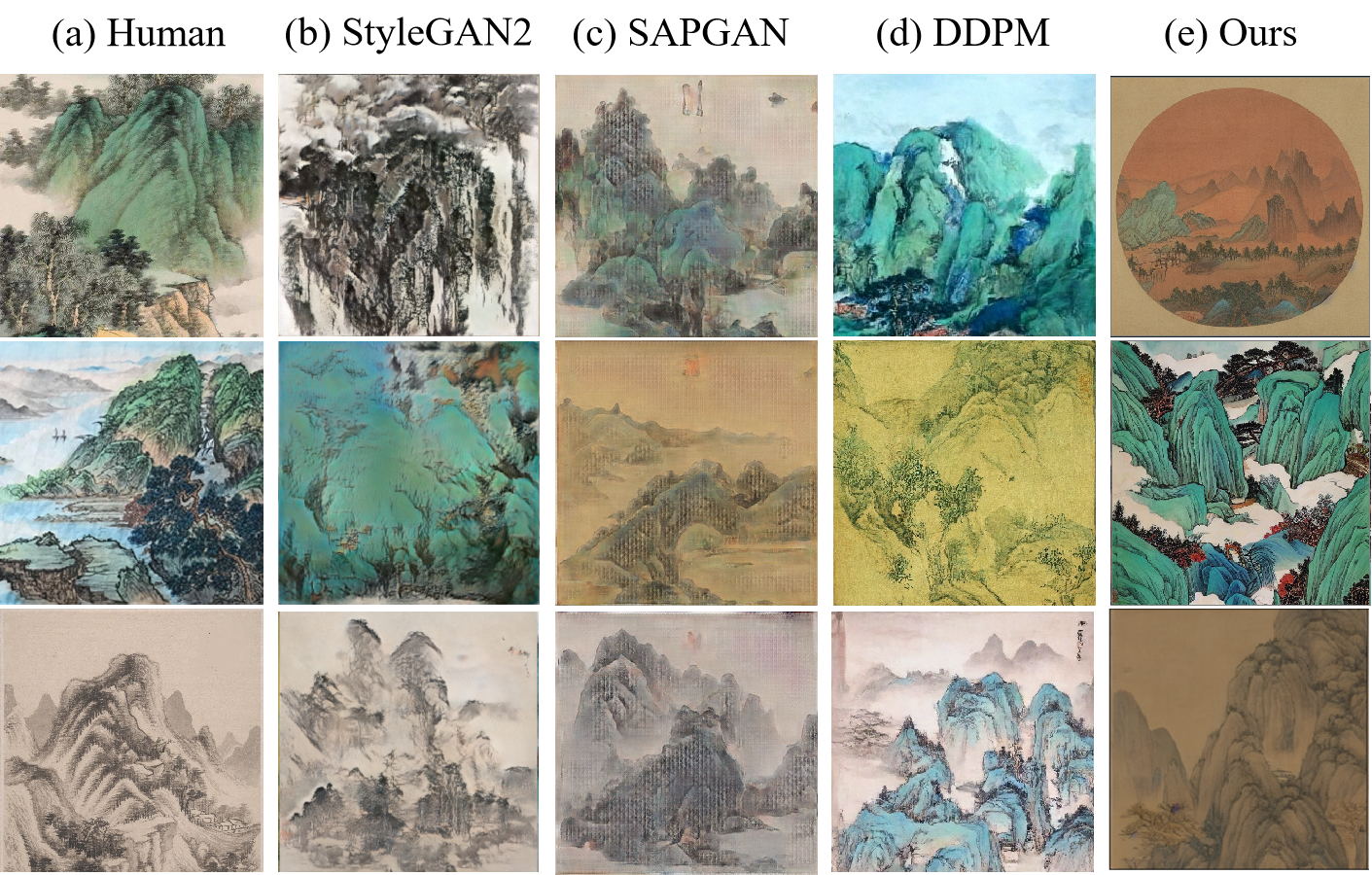}}
    \caption{Qualitative Comparison between Chinese landscape paintings generated by (a) Human, (b) StyleGAN2, (c) SAPGAN, (d) DDPM, and (e) Ours. Specifically, the input text of our model is “a Chinese landscape painting”.}
    \label{fig:comparison}
\end{figure}

\subsection{Style Aggregator}

In order to control the style of the generated paintings, we infuse the user-given style into the paintings generated by the Content Generator. Here, we utilize PAMA \cite{Luo_2022_ACCV}, a state-of-art arbitrary style transfer method, to construct style aggregator  $\mathcal{S}$. To be specific, it gradually aligns content manifolds with style manifolds through an attention mechanism to ensure consistent stylization between semantic regions. Compared with other models, PAMA can better maintain consistency of semantic regions, which allows the style of the reference painting to be maintained in the stylized painting.

Given a generated painting $I_{Content}$ and a reference painting $I_{Style}$, a pre-trained VGG19 network $\mathcal{M}$ encodes them into features $F_{Content}$ and $F_{Style}$, respectively. Then, through three attentional manifold alignment (AMA) block, content feature $F_{Content}$ will be gradually integrated with style information $F_{Style}$ and then be decoded to stylized painting $I_{Output}$ through $\mathcal{N}$ whose structure is symmetric to the encoder $\mathcal{M}$. The loss function can be summarized as:
\begin{equation}
    \mathcal{L_S} = \sum_{i=1}(\lambda_{ss}^i L_{ss} + \lambda_{r}^i L_{r} + \lambda_{m}^i L_{m} + \lambda_{h}^i L_{h}) + L_{rec}
\end{equation}
Where $L_{ss}$ is the content loss for content preserving, $L_r$, $L_m$, and $L_h$ are the style losses for maintaining the certain style and $L_{rec}$ refers to the image reconstruction loss, $i$ denotes the i-th AMA and $\lambda_{x}^i$ are the weight for $L_{x}$.

\begin{figure*}[htbp]
    \centerline{\includegraphics[scale=0.54]{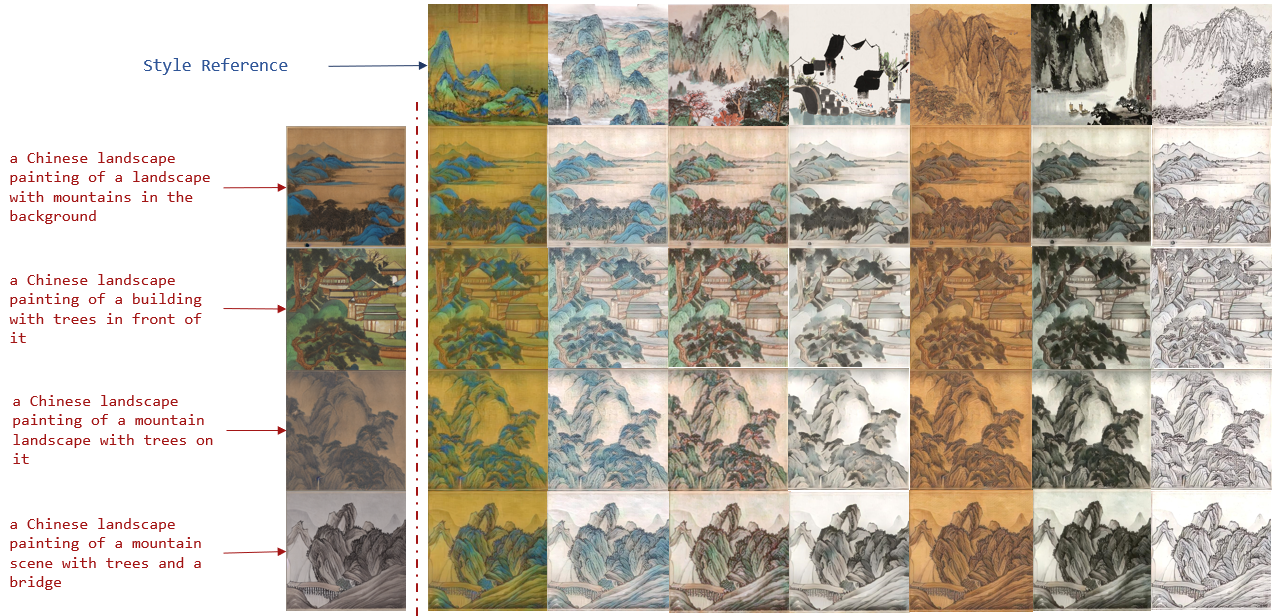}}
    \caption{Samples of our controllable generation paintings. The first row displays the style reference painting, while the red texts correspond to the semantic information of generated paintings. }
    \label{fig:final result}
\end{figure*}

\begin{figure}[htbp]
    \centerline{\includegraphics[scale=0.67]{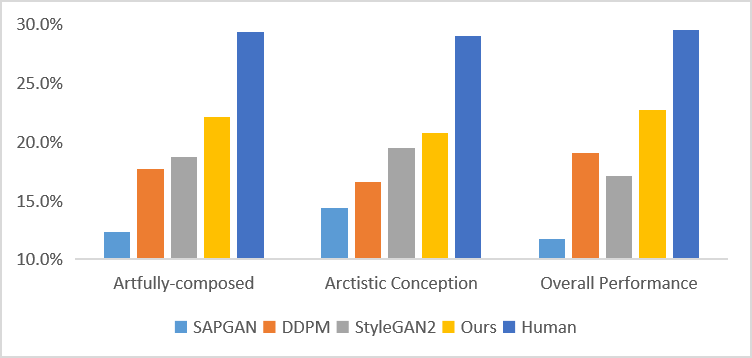}}
    \caption{User preference result of all methods. The higher, the better.}
    \label{fig:AA}
\end{figure}

\begin{figure}[htbp]
    \centerline{\includegraphics[scale=0.67]{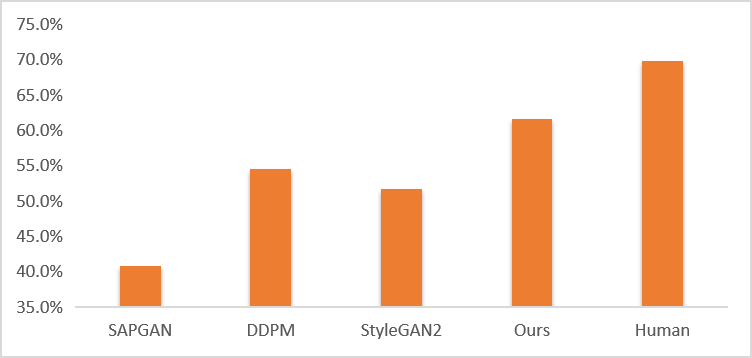}}
    \caption{Result on Visual Turing Test, asking participants to judge if an artwork is made by a human or computer. The higher, the better.}
    \label{fig:BA}
\end{figure}

\section{Experiments}

In this section, we firstly introduce the implementation details of our method. Then we compare our CCLAP with the state-of-the-art methods in both qualitative and quantitative evaluation. Finally, we give more visualization results of our method for controllable Chinese landscape painting generation.

\subsection{Implementation Details}

We fine-tune our Content Generator, beginning with the model trained on LAION-5B \cite{Schuhmann2022LAION5BAO}, and run training for around 35 epochs with a batch size of 4 on our proposed CLAP dataset. We fine-tune the Style Aggregator on our proposed dataset, which is initialized by a pre-trained model trained on WikiArt \cite{wikiart} and MS-COCO \cite{Lin2014MicrosoftCC}. We run training with a batch size of 4 for 60000 iterations.  The weights $\lambda_{ss}^1, \lambda_{ss}^2, \lambda_{ss}^3$ are set to 12, 9, 7, respectively, while $\lambda_{h}^1, \lambda_{h}^2, \lambda_{h}^3$ are set to 0.25, 0.5, and 1, respectively. All the weights $\lambda_{r}^i$, $\lambda_{m}^i$  of $L_{r}$ and $L_{m}$ are set to 2.

For fair comparisons, we re-train DDPM \cite{NEURIPS2020_4c5bcfec}, SAPGAN \cite{Xue2020EndtoEndCL}, and StyleGAN2 \cite{Karras2019AnalyzingAI} on the same dataset CLAP. It should be noted that there is no public code for SAPGAN, thus we reproduce it following the settings mentioned in the original article \cite{Xue2020EndtoEndCL}. 

\subsection{Comparisons With State-of-the-arts}

Since almost all existing methods can not conduct controllable Chinese landscape painting generation, we simplify our CLAP to randomly generate paintings by taking ``a Chinese landscape painting'' as input and then compare with current the-state-of-arts methods. 

\subsubsection{Qualitative Evaluation}

To verify the effectiveness of our CCLAP, we compare our method with StyleGAN2 \cite{Karras2019AnalyzingAI} , SAPGAN \cite{Xue2020EndtoEndCL}, and DDPM \cite{NEURIPS2020_4c5bcfec} qualitatively. As shown in Fig.\ref{fig:comparison},  SAPGAN \cite{Xue2020EndtoEndCL} decomposes the generation into two steps, where the first stage produces an outline without semantic details. Thus, it tends to neglect the local details, such as trees and the texture of mountains as shown in the second and third rows. StyleGAN2 \cite{Karras2019AnalyzingAI} gets better results in generating mountains as there are clear textures on the mountains. However, their artistic composition needs to be further improved. For example, in the first and second rows, the drawings with similar contents are full of the whole painting, which results in lower fidelity. DDPM \cite{NEURIPS2020_4c5bcfec} can generate more elements in the painting, e.g., clouds and trees. Our method outperforms all these existing methods with better diversity, reflected in the generated landscape paintings that contain richer and finer details, as well as more varied artistic compositions, as shown in the first and second rows.

\subsubsection{Quantitative Evaluation}

Furthermore, we resort to user study to conduct quantitative evaluations. Specifically, we generate 100 paintings and randomly select 10 of them for each method. Then all 50 paintings (including 10 human paintings) are divided into ten groups randomly and then distributed to users. For each group, users are asked the following question:

\textbf{Q:} Please choose your favorite painting in terms of artfully-composed, artistic conception, and overall performance separately.

We choose the above three evaluation metrics as they give a comprehensive quantitative evaluation of paintings, from low-level drawing elements to high-level intuitive feeling. We recruit a total of 70 volunteers to participate the Test. 

\textbf{Results.} Fig.\ref{fig:AA}. shows users' preference results of all methods. Our method gets the best results in all three metrics compared with other generative models. Although our method achieves promising results, it is still much worse than humans. Both the artfully-composed and artistic conception should be further improved, which will be explored in the future by explicitly inducing human painting knowledge into model design.

Furthermore, we make a simplified Turing Test by asking volunteers to judge if the painting is made by human beings. The proportion $p$ of “real paintings” is calculated by the following equation.
\begin{equation}
    p = \frac{N_r}{N_v * N_i}
\end{equation}
where $N_r$ means the total number of paintings selected to be real for each method by $N_v$ volunteers. $N_i$ denotes the number of paintings generated by each method for the Test. In our experiments, $N_v$=70, $N_i$=10. As shown in Fig.\ref{fig:BA}, our paintings are judged as human art with a proportion of 61.7$\%$, which is much higher than all other methods. Interestingly, only 69.6$\%$ of human paintings are judged as real, demonstrating that our method can generate promising results in terms of reality. 

\subsection{Controllable Painting Generation}

Since our CCLAP is a controllable Chinese landscape painting method that can generate the paintings specific to the text input and the style indicated in the reference image, we conduct experiments on generating paintings with different texts or style reference images as input. As shown in Fig.\ref{fig:final result}, 28 paintings are generated by taking four different texts and seven distinct style reference images. CCLAP can well generate buildings or bridges according to the input text descriptions. 
Moreover, we choose representative styles as reference images to generate different paintings, as shown in the row direction of Fig.\ref{fig:final result}. All these results demonstrate that our method can conduct promising controllable Chinese landscape paintings by taking texts or different styles reference images as inputs.

\subsection{Some Failure Cases}
\begin{figure}[tbp]
    \centerline{\includegraphics[scale=0.45]{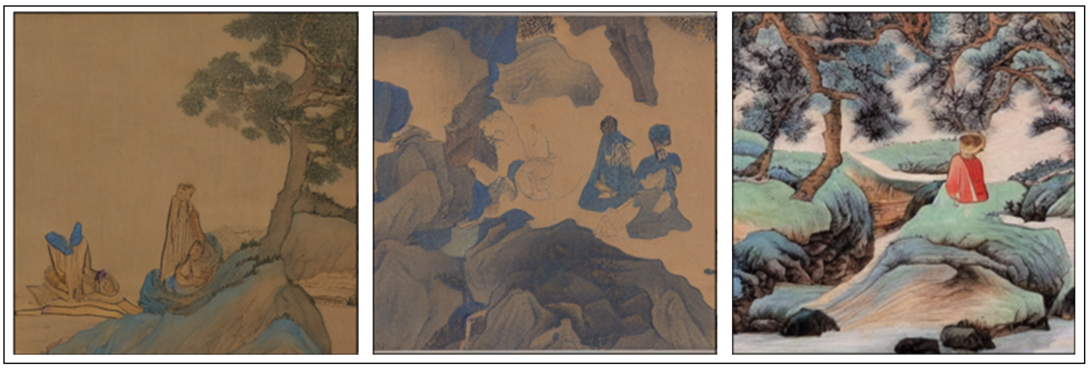}}
    \caption{Some failure cases of our CCLAP. Input: a Chinese landscape painting of a \textbf{man} standing on top of a cliff.}
    \label{f1}
\end{figure}

Fig.\ref{f1} shows some failure cases of our CCLAP. As seen, our model can not well generate human beings at present. We believe that there are two reasons behind. On the one hand, the training data containing the keyword "man" is relatively small, accounting for less than 10$\%$; On the other hand, the sizes of human beings take up a small proportion in the whole landscape painting (usually less than 5$\%$), which may be neglected during generation. Possible solutions to this problem could be to increase the paintings containing human beings and design a special architecture with attention mechanism for better perceiving and modeling human beings. We will investigate these possibilities in the future.

\section{Conclusion}

This paper proposes CCLAP to generate controllable Chinese landscape paintings based on Latent Diffusion Model by taking texts and different reference images as inputs. Our method consists of two key modules, i.e., the content generator and the style aggregator. The content generator can achieve paintings of the specific content to the input text while the style aggregator is conducted to generate various styles of paintings specific to the reference image. Moreover, to facilitate the experiments, we built a new dataset called CLAP which consists of 3560 samples paired with the painting and corresponding text descriptions. Extensive experiments show that our CCLAP outperforms the state-of-the-art performance in terms of both qualitative  and quantitative comparisons.

\bibliographystyle{IEEEbib}
{
\bibliography{icme2023template}}

\begin{thebibliography}{10}

\bibitem{Xue2020EndtoEndCL}
Alice Xue,
\newblock ``End-to-end chinese landscape painting creation using generative
  adversarial networks,''
\newblock in {\em the IEEE Winter Conference on Applications of Computer Vision
  (WACV)}, 2020.

\bibitem{Luo2022HighResolutionAA}
Pei Luo, Jinchao Zhang, and Jie Zhou,
\newblock ``High-resolution and arbitrary-sized chinese landscape painting
  creation based on generative adversarial networks,''
\newblock in {\em the Thirty-First International Joint Conference on Artificial
  Intelligence (IJCAI)}, 2022.

\bibitem{9903093}
Yongxing He, Wei Li, Z.~Li, and Yongchuan Tang,
\newblock ``Gluegan: Gluing two images as a panorama with adversarial
  learning,''
\newblock {\em 2022 14th International Conference on Intelligent Human-Machine
  Systems and Cybernetics (IHMSC)}, pp. 196--201, 2022.

\bibitem{Zhou2019AnIA}
Le~Zhou, Qiu-Feng Wang, Kaizhu Huang, and Cheng-Hung Lo,
\newblock ``An interactive and generative approach for chinese shanshui
  painting document,''
\newblock in {\em the International Conference on Document Analysis and
  Recognition (ICDAR)}, 2019, pp. 819--824.

\bibitem{He2018ChipGANAG}
Bin He, Feng Gao, Daiqian Ma, Boxin Shi, and Ling yu~Duan,
\newblock ``Chipgan: A generative adversarial network for chinese ink wash
  painting style transfer,''
\newblock {\em Proceedings of the 26th ACM international conference on
  Multimedia (ACM MM)}, 2018.

\bibitem{Lin2018TransformAS}
Daoyu Lin, Yang Wang, Guangluan Xu, Jun~Yu Li, and Kun Fu,
\newblock ``Transform a simple sketch to a chinese painting by a multiscale
  deep neural network,''
\newblock {\em Algorithms}, vol. 11, pp. 4, 2018.

\bibitem{Yuan2022LearningTG}
Shaozu Yuan, Aijun Dai, Zhiling Yan, Ruixue Liu, Meng Chen, Baoyang Chen,
  Zhijie Qiu, and Xiaodong He,
\newblock ``Learning to generate poetic chinese landscape painting with
  calligraphy,''
\newblock in {\em the Thirty-First International Joint Conference on Artificial
  Intelligence (IJCAI)}, 2022.

\bibitem{Lv2019GeneratingCC}
Xia Lv and Xiwen Zhang,
\newblock ``Generating chinese classical landscape paintings based on
  cycle-consistent adversarial networks,''
\newblock {\em 2019 6th International Conference on Systems and Informatics
  (ICSAI)}, pp. 1265--1269, 2019.

\bibitem{Goodfellow2014GenerativeAN}
Ian~J. Goodfellow, Jean Pouget-Abadie, Mehdi Mirza, Bing Xu, David
  Warde-Farley, Sherjil Ozair, Aaron~C. Courville, and Yoshua Bengio,
\newblock ``Generative adversarial nets,''
\newblock in {\em Proceedings of the Advances in Neural Information Processing
  Systems (NIPS)}, 2014.

\bibitem{10.5555/3157096.3157346}
Tim Salimans, Ian Goodfellow, Wojciech Zaremba, Vicki Cheung, Alec Radford, and
  Xi~Chen,
\newblock ``Improved techniques for training gans,''
\newblock in {\em Proceedings of the Advances in Neural Information Processing
  Systems (NIPS)}, 2016.

\bibitem{NEURIPS2020_4c5bcfec}
Jonathan Ho, Ajay Jain, and Pieter Abbeel,
\newblock ``Denoising diffusion probabilistic models,''
\newblock in {\em Advances in Neural Information Processing Systems},
  H.~Larochelle, M.~Ranzato, R.~Hadsell, M.F. Balcan, and H.~Lin, Eds. 2020,
  vol.~33, pp. 6840--6851, Curran Associates, Inc.

\bibitem{Rombach2021HighResolutionIS}
Robin Rombach, A.~Blattmann, Dominik Lorenz, Patrick Esser, and Bj{\"o}rn
  Ommer,
\newblock ``High-resolution image synthesis with latent diffusion models,''
\newblock in {\em Proceedings of the IEEE/CVF Conference on Computer Vision and
  Pattern Recognition (CVPR)}, 2021, pp. 10674--10685.

\bibitem{Luo_2022_ACCV}
Xuan Luo, Zhen Han, and Linkang Yang,
\newblock ``Progressive attentional manifold alignment for arbitrary style
  transfer,''
\newblock in {\em Asian Conference on Computer Vision (ACCV)}, December 2022,
  pp. 3206--3222.

\bibitem{Karras2019AnalyzingAI}
Tero Karras, Samuli Laine, Miika Aittala, Janne Hellsten, Jaakko Lehtinen, and
  Timo Aila,
\newblock ``Analyzing and improving the image quality of stylegan,''
\newblock in {\em Proceedings of the IEEE/CVF Conference on Computer Vision and
  Pattern Recognition (CVPR)}, 2019, pp. 8107--8116.

\bibitem{Wang2018ESRGANES}
Xintao Wang, Ke~Yu, Shixiang Wu, Jinjin Gu, Yihao Liu, Chao Dong, Chen~Change
  Loy, Yu~Qiao, and Xiaoou Tang,
\newblock ``Esrgan: Enhanced super-resolution generative adversarial
  networks,''
\newblock in {\em Proceedings of the European Conference on Computer Vision
  (ECCV) Workshops}, 2018.

\bibitem{Li2018NeuralAS}
B.~Li, Caiming Xiong, Tianfu Wu, Yu~Zhou, Lun Zhang, and Rufeng Chu,
\newblock ``Neural abstract style transfer for chinese traditional painting,''
\newblock pp. 212--227, 2018.

\bibitem{Wang2021AttentionalWN}
Rui Wang, Huaibo Huang, Aihua Zheng, and Ran He,
\newblock ``Attentional wavelet network for traditional chinese painting
  transfer,''
\newblock in {\em Proceedings of the 25th International Conference on Pattern
  Recognition (ICPR)}, 2021, pp. 3077--3083.

\bibitem{Radford2015UnsupervisedRL}
Alec Radford, Luke Metz, and Soumith Chintala,
\newblock ``Unsupervised representation learning with deep convolutional
  generative adversarial networks,''
\newblock {\em CoRR}, vol. abs/1511.06434, 2015.

\bibitem{Arjovsky2017WassersteinG}
Mart{\'i}n Arjovsky, Soumith Chintala, and L{\'e}on Bottou,
\newblock ``Wasserstein gan,''
\newblock p. 214–223, 2017.

\bibitem{Ramesh2022HierarchicalTI}
Aditya Ramesh, Prafulla Dhariwal, Alex Nichol, Casey Chu, and Mark Chen,
\newblock ``Hierarchical text-conditional image generation with clip latents,''
\newblock 2022, vol. abs/2204.06125.

\bibitem{liao2020real}
Minghui Liao, Zhaoyi Wan, Cong Yao, Kai Chen, and Xiang Bai,
\newblock ``Real-time scene text detection with differentiable binarization,''
\newblock in {\em AAAI Conference on Artificial Intelligence(AAAI)}, 2020.

\bibitem{Telea2004AnII}
Alexandru~Cristian Telea,
\newblock ``An image inpainting technique based on the fast marching method,''
\newblock {\em Journal of Graphics Tools}, vol. 9, pp. 23 -- 34, 2004.

\bibitem{Li2022BLIPBL}
Junnan Li, Dongxu Li, Caiming Xiong, and Steven C.~H. Hoi,
\newblock ``Blip: Bootstrapping language-image pre-training for unified
  vision-language understanding and generation,''
\newblock in {\em Proceedings of the 34th International Conference on Machine
  Learning (ICML)}, 2022.

\bibitem{Schuhmann2022LAION5BAO}
Christoph Schuhmann, Romain Beaumont, Richard Vencu, Cade Gordon, Ross
  Wightman, Mehdi Cherti, Theo Coombes, Aarush Katta, Clayton Mullis, Mitchell
  Wortsman, Patrick Schramowski, Srivatsa Kundurthy, Katherine Crowson, Ludwig
  Schmidt, Robert Kaczmarczyk, and Jenia Jitsev,
\newblock ``Laion-5b: An open large-scale dataset for training next generation
  image-text models,''
\newblock {\em ArXiv}, vol. abs/2210.08402, 2022.

\bibitem{wikiart}
K.~Nichol,
\newblock ``Painter by numbers,'' 2016.

\bibitem{Lin2014MicrosoftCC}
Tsung-Yi Lin, Michael Maire, Serge~J. Belongie, James Hays, Pietro Perona, Deva
  Ramanan, Piotr Doll{\'a}r, and C.~Lawrence Zitnick,
\newblock ``Microsoft coco: Common objects in context,''
\newblock in {\em Proceedings of the European Conference on Computer Vision
  (ECCV)}, 2014.

\end{thebibliography}

\onecolumn
\begin{appendix}
\subsection{More Results with a higher resolution}

Here we give more results of our CCLAP with different texts as inputs. As seen, CCLAP can well generate high-quality Chinese landscape painting under various conditions. It should be noted that the most important and common elements in Chinese landscape paintings are landscapes, rivers, trees and buildings. Our method can generate diverse paintings in terms of these elements specific to the input text.

\begin{figure}[!h]
    \centering
    \includegraphics[scale=0.7]{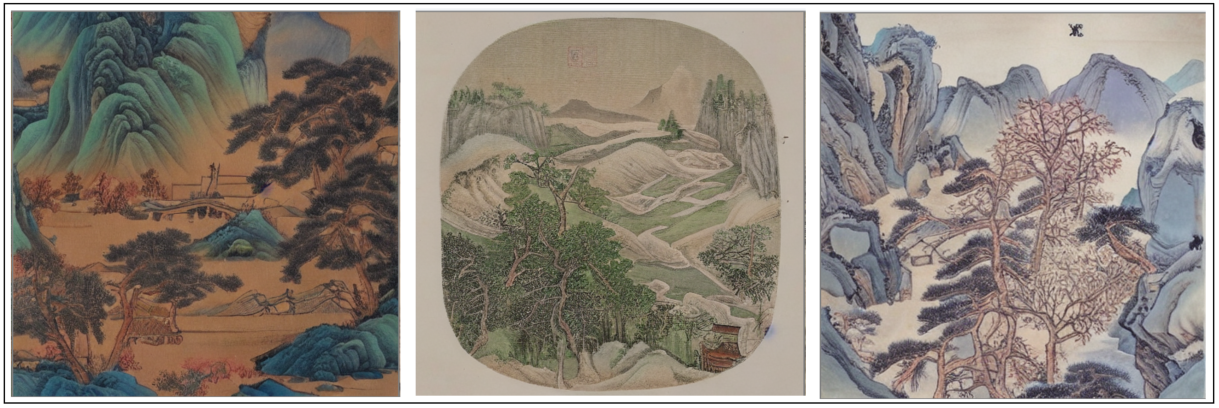}
    \caption{A Chinese landscape painting of a \textbf{mountain} landscape with \textbf{trees}}
\end{figure}
\begin{figure}[!h]
    \centering
    \includegraphics[scale=0.7]{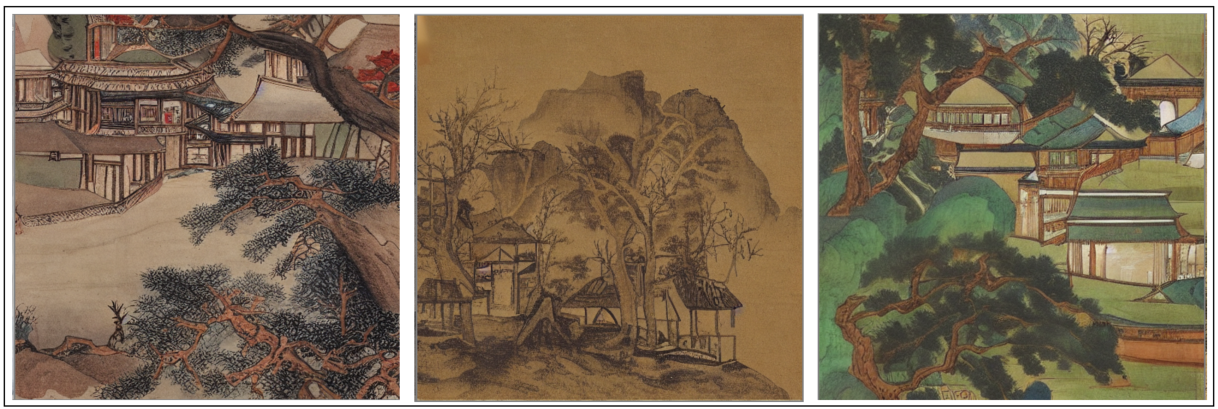}
    \caption{A Chinese landscape painting of a \textbf{building} with \textbf{trees} in front of it}
\end{figure}
\begin{figure}[!h]
    \centering
    \includegraphics[scale=0.7]{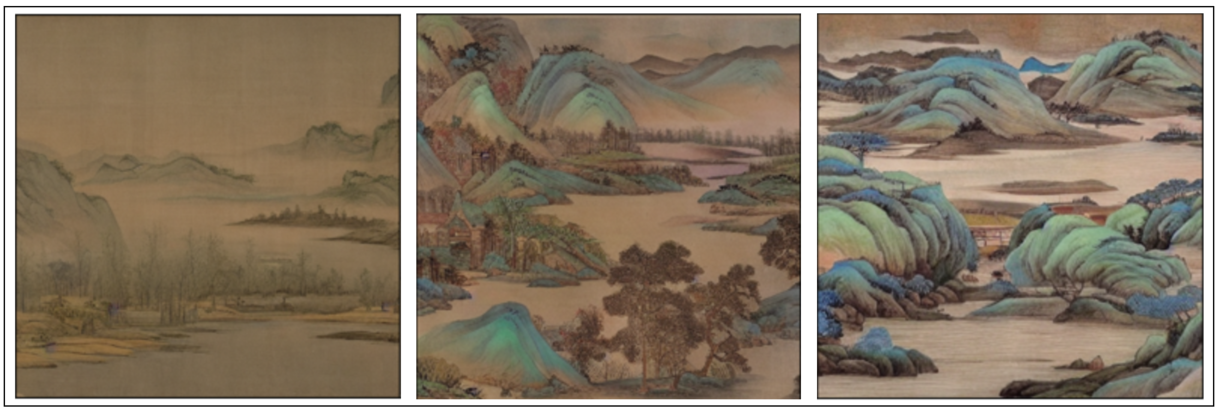}
    \caption{A Chinese landscape painting of a landscape with \textbf{mountains} and a \textbf{river}}
\end{figure}
\begin{figure}[!h]
    \centering
    \includegraphics[scale=0.7]{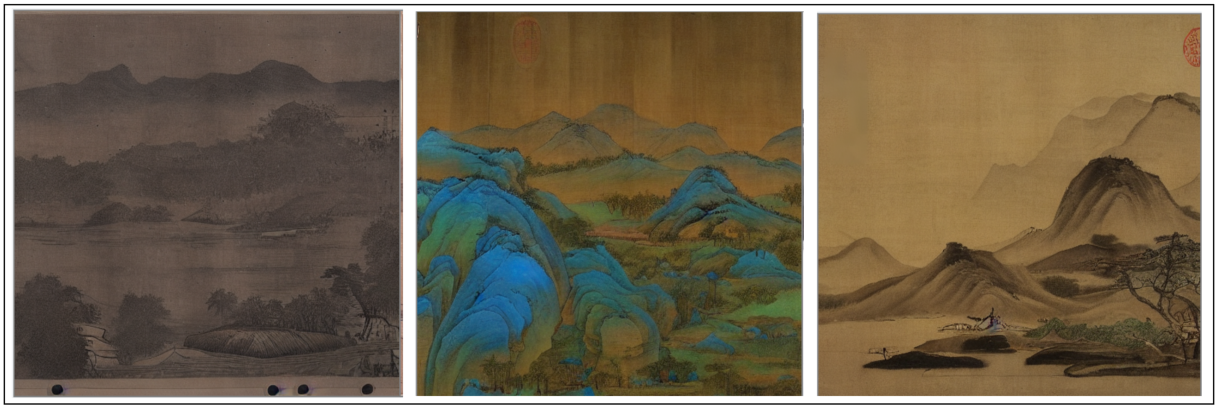}
    \caption{A Chinese landscape painting of a landscape with \textbf{mountains} in the background}
\end{figure}
\begin{figure}[!h]
    \centering
    \includegraphics[scale=0.78]{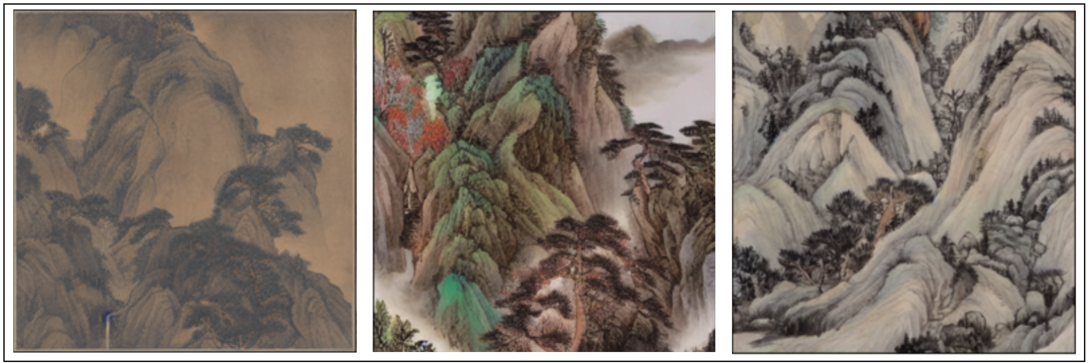}
    \caption{A Chinese landscape painting of a \textbf{mountain} landscape with \textbf{trees} on it}
\end{figure}
\begin{figure}[!h]
    \centering
    \includegraphics[scale=0.7]{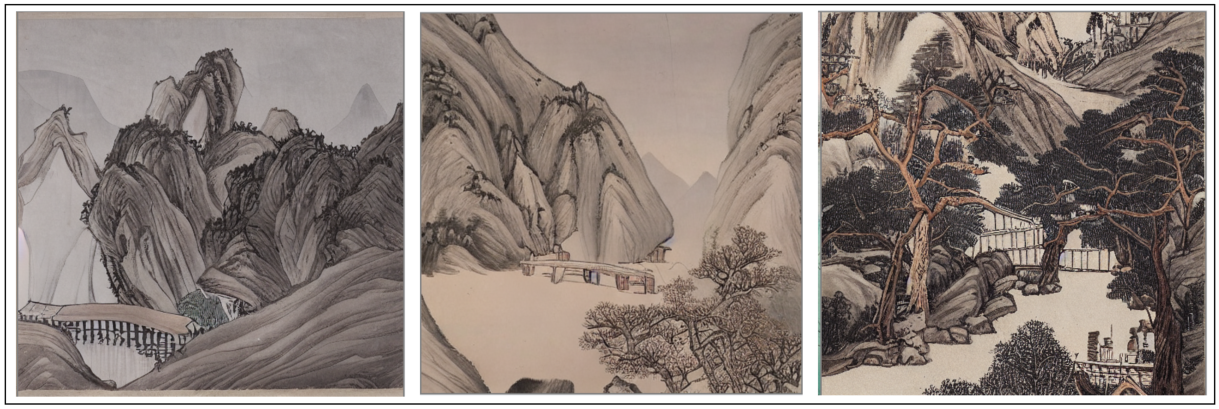}
    \caption{A Chinese landscape painting of a mountain scene with \textbf{trees} and a \textbf{bridge}}
\end{figure}

\end{appendix}

\end{document}